\documentclass[conference]{IEEEtran}
\IEEEoverridecommandlockouts

\usepackage{cite}
\usepackage{amsmath,amssymb,amsfonts}
\usepackage{algorithmic}
\usepackage{url}
\usepackage{graphicx}
\usepackage{textcomp}
\usepackage{threeparttable}
\usepackage{titlesec}
\usepackage{booktabs}
\usepackage[caption=false,font=footnotesize]{subfig}
\usepackage[letterpaper,top=54pt,bottom=54pt,left=54pt,right=54pt]{geometry}
\usepackage{stfloats}
\usepackage{xcolor}

\def\BibTeX{{\rm B\kern-.05em{\sc i\kern-.025em b}\kern-.08em
    T\kern-.1667em\lower.7ex\hbox{E}\kern-.125emX}}

\titlespacing*{\section}{0pt}{0.8ex plus 0.2ex}{0.3ex plus 0.1ex}
\titlespacing*{\subsection}{0pt}{0.6ex plus 0.2ex}{0.2ex plus 0.1ex}
\titlespacing*{\subsubsection}{0pt}{0.4ex plus 0.1ex}{0.1ex plus 0.1ex}

\usepackage{etoolbox}

\makeatletter
\patchcmd{\@maketitle}
  {\newpage}
  {\vspace*{18pt}\newpage}  
  {}{}
\makeatother

\begin{document}

\title{Eyes on the Road: A Naturalistic Comparison of MTW Rider Gaze in Urban Indian Traffic\\
}

\author{\IEEEauthorblockN{1\textsuperscript{st} Prerak Srivastava}
\IEEEauthorblockA{
\textit{IIIT Hyderabad}\\
Hyderabad, India \\
prerak.s@research.iiit.ac.in}
\and
\IEEEauthorblockN{2\textsuperscript{nd} Bhaiya Vaibhaw Kumar}
\IEEEauthorblockA{
\textit{IIIT Hyderabad}\\
Hyderabad, India \\
bhaiya.vaibhaw@research.iiit.ac.in
}
\and
\IEEEauthorblockN{1\textsuperscript{st} Kavita Vemuri}
\IEEEauthorblockA{
\textit{IIIT Hyderabad}\\
Hyderabad, India \\
kvemuri@iiit.ac.in
}
}

\maketitle

\begin{abstract}
Motorized two-wheelers (MTW) dominate Indian roads but remain underrepresented in driver behavior research. This study presents the first large-scale analysis of MTW driver gaze behavior in naturalistic, heterogeneous urban traffic, using the \textit{myEye2Wheeler} dataset. A semantic segmentation pipeline (YOLOv11 + SAM2) was used to extract object-level gaze metrics under two attention modes: direct gaze (foveal overlap) and central vision (parafoveal monitoring). Results reveal a functional division: central vision supports broad monitoring, while direct gaze enables brief, selective sampling. Novice riders exhibit road-anchored scanning, returning to the road between object fixations, while experienced riders form longer chains of attention across multiple objects. The findings suggest that experience primarily refines temporal rhythm rather than altering allocation strategy and reduces object-class effects in gaze patterns. These findings offer new insight into MTW attention structures and inform future work on behavior modeling and safety systems.
\end{abstract}

\vspace{-10pt}
\section{Introduction}
According to the Indian Ministry of Road Transport and Highways, there are around 260 million motorized two-wheeler vehicles (MTW) on Indian roads, with around 185 MTWs and 34 cars per 1000 people \cite{DataForIndia2024}. Navigation strategies of MTW drivers differ from those of four-wheeler drivers due to their smaller size and greater agility, enabling flexible movement through dense traffic \cite{kumar2025myeye2wheeler}. Traffic complexity arises from the absence of dedicated lanes for MTWs, resulting in prevalent non-lane adherence behavior.

MTW drivers require heightened visual awareness due to greater exposure to environmental factors like weather, road surface variations and surrounding vehicles \cite{wigum2023analysis}. As a result, existing driver behavior research developed on data from four-wheeler drivers does not adequately capture MTW behavior in heterogeneous traffic, highlighting the need for focused MTW-specific research.

This study conducts a comparative eye-gaze analysis of novice and experienced MTW drivers using the \textit{myEye2Wheeler} dataset \cite{kumar2025myeye2wheeler}. We believe that our findings would assist in developing vehicle-type-focused road and traffic management strategies, improving driver training techniques, and contributing to traffic safety research.

\subsection{Research Gaps}

Despite growing interest in driver behavior analysis, several major gaps remain: limited studies conducted on MTW drivers and inadequate research done on naturalistic driver behavior datasets. Most attention-related studies focus on four-wheeler drivers, with few studies investigating MTW riders, particularly in non-Western traffic environments \cite{DiStasi2011, Papakostopoulos2020}. Moreover, naturalistic studies on gaze behavior of MTW are scarce and often restricted by small sample sizes, limiting the generalizability of findings on attention allocation and decision-making patterns \cite{aupetit2012naturalistic, deniaud2015presence}. This study addresses a few gaps by developing a comprehensive data processing pipeline using semantic image segmentation to efficiently process naturalistic driver gaze data, coupled with the application of rigorous statistical methods for analysis and derive structured inferences based on real-life driver behavior.

\subsection{Related Works}\subsubsection{Naturalistic Eye-Tracking Usage in Driver Research}
Given that driving is primarily a visuo-motor task \cite{Kotseruba2021}, studies have established that fixation based eye-tracking metrics correlate strongly with driver situation awareness \cite{Portela2024}. Eye-tracking technique has significantly advanced driver behavior research as it provides a direct and precise method of capturing when, where, and for how long drivers focus their gaze \cite{Ahlstrom2021}. Its strength lies in the detailed oculomotor metrics it provides, most notably \textit{scanpaths} - the visual trace of the eye movements over time from which we  obtain \textit{saccades} - rapid gaze movements between fixations characterized by amplitude, velocity and frequency and \textit{fixations}, which are periods of gaze stability on Areas of Interest (AOIs) quantified using count and duration \cite{holmqvist2011}.

Similar to Western datasets like \textit{DR(Eye)VE} \cite{Palazzi2019} and BDD-A \cite{Xia2018}, which capture driver gaze in structured Western traffic environments, Indian datasets such as IDD \cite{varma2019idd} and DATS\_2022 \cite{paranjape2022dats} focus on unstructured Indian traffic but they either lack gaze data or representation of MTW drivers. Addressing the gap in datasets collected in naturalistic settings, the \textit{myEye2Wheeler} dataset captures synchronized eye-tracking and scene video data from MTW drivers navigating unstructured urban traffic \cite{kumar2025myeye2wheeler}.

\subsubsection{Factors Affecting Driver Gaze Behavior}

Driving experience significantly affects visual attention allocation and scanning patterns, which is reflected in the efficacy of visual exploration \cite{Jackson2009}. Studies \cite{Underwood2011, Underwood2003} show that experienced drivers are able to adapt visual search strategies as a function of driving complexity, showing greater sensitivity to road types and potential hazards compared to novice drivers. They also exhibit a wider horizontal gaze spread \cite{Konstantopoulos2010} and tend to scan scenes more thoroughly with shorter mean saccades \cite{Kotseruba2021}.

Attentional strategies in driving involve both gaze distribution and the influence of cognitive load (e.g., experienced versus novice drivers) on scanning behavior, with peripheral vision \cite{WOLFE2017} playing a key role in guiding attention. Experienced drivers often detect hazards without directly fixating on them, using brief glances to monitor road elements in their periphery, a pattern also observed in simulated experiments\cite{Underwood2011, Underwood2003}. Moreover, fixation-centric metrics alone may misrepresent driver attention, highlighting the need to quantify how peripheral information supports decision-making \cite{Ahlstrom2021}. The type of vehicle also influences the patterns of visual attention. Naturalistic study comparing motorcyclists and automobile drivers found similar fixation patterns on highways but in urban traffic, motorcyclists demonstrated nearly twice as many fixations per unit time as compared to automobile drivers, directed at potential hazards \cite{Papakostopoulos2010}.

\section{Methodology}

\subsection{Data Preprocessing and Selection}

The \textit{myEye2Wheeler} dataset contains 40 participants egocentric videos with eye tracking (Tobii Glasses 2) captured from MTW riders driving in urban Indian traffic, at 1920$\times$1080 pixel resolution and 25 fps. The drivers were classified into two groups: novice and experienced drivers \cite{kumar2025myeye2wheeler}. As a preprocessing step, videos that were too short or had high levels of scene or eye-tracking noise were excluded from the analysis, primarily due to the drivers' excessive head movement. To ensure that both driver groups had an equal number of frames for analysis, the remaining videos were screened based on the quality of the gaze data. Each frame was assessed for the presence of a localized red circle (driver's gaze position), extracted using contour-based image processing. Videos with the highest gaze coverage were selected while keeping the frame counts between the two groups roughly balanced. This process yielded a final dataset of 15 experienced and 10 novice drivers, contributing 109{,}339 and 107{,}673 frames with an average gaze coverage of 85.86\% and 87.26\%, respectively.

\subsection{Semantic Image Segmentation Pipeline for Traffic Scene Analysis}
\begin{figure}[htbp]
\centering
\subfloat[\label{fig:pipeline_demo}]{
    \includegraphics[width=0.47\linewidth]{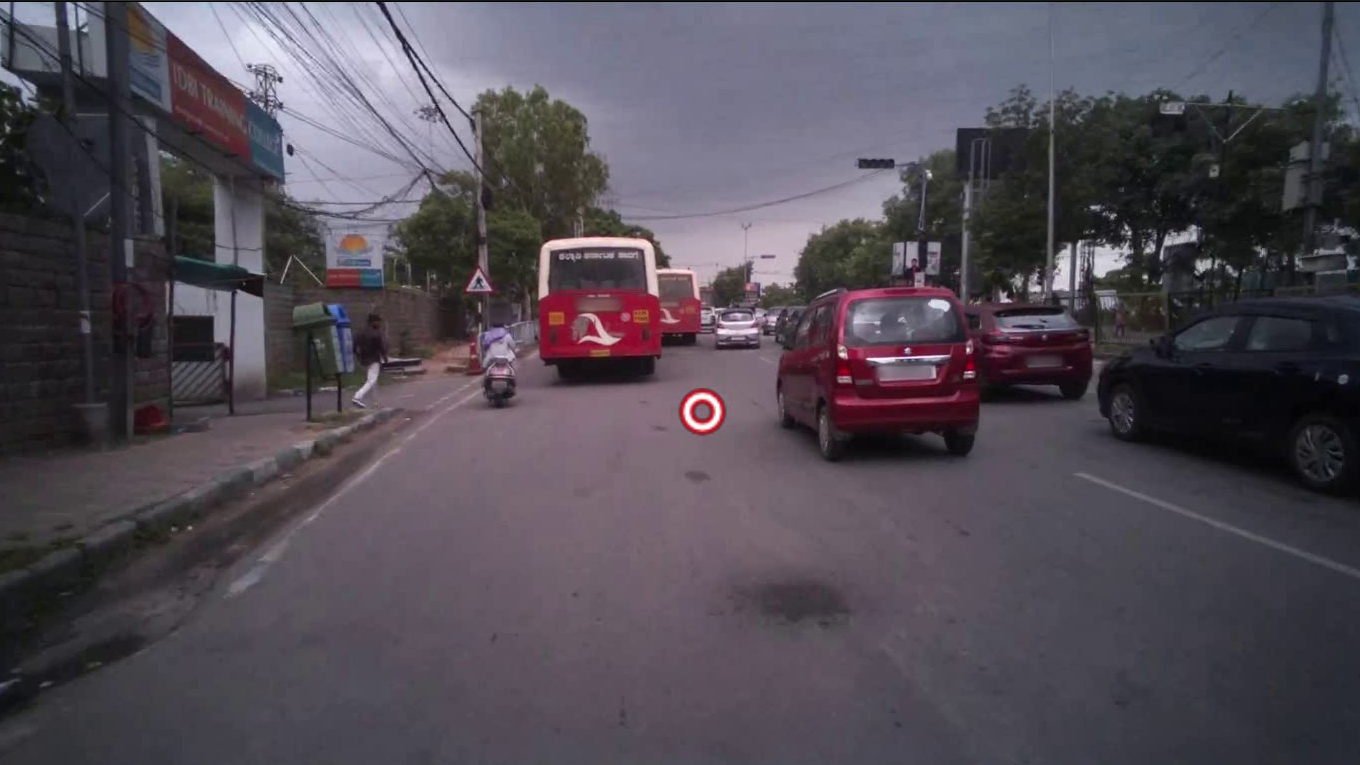}
}
\hfill
\subfloat[\label{fig:pipeline_viz}]{
    \includegraphics[width=0.47\linewidth]{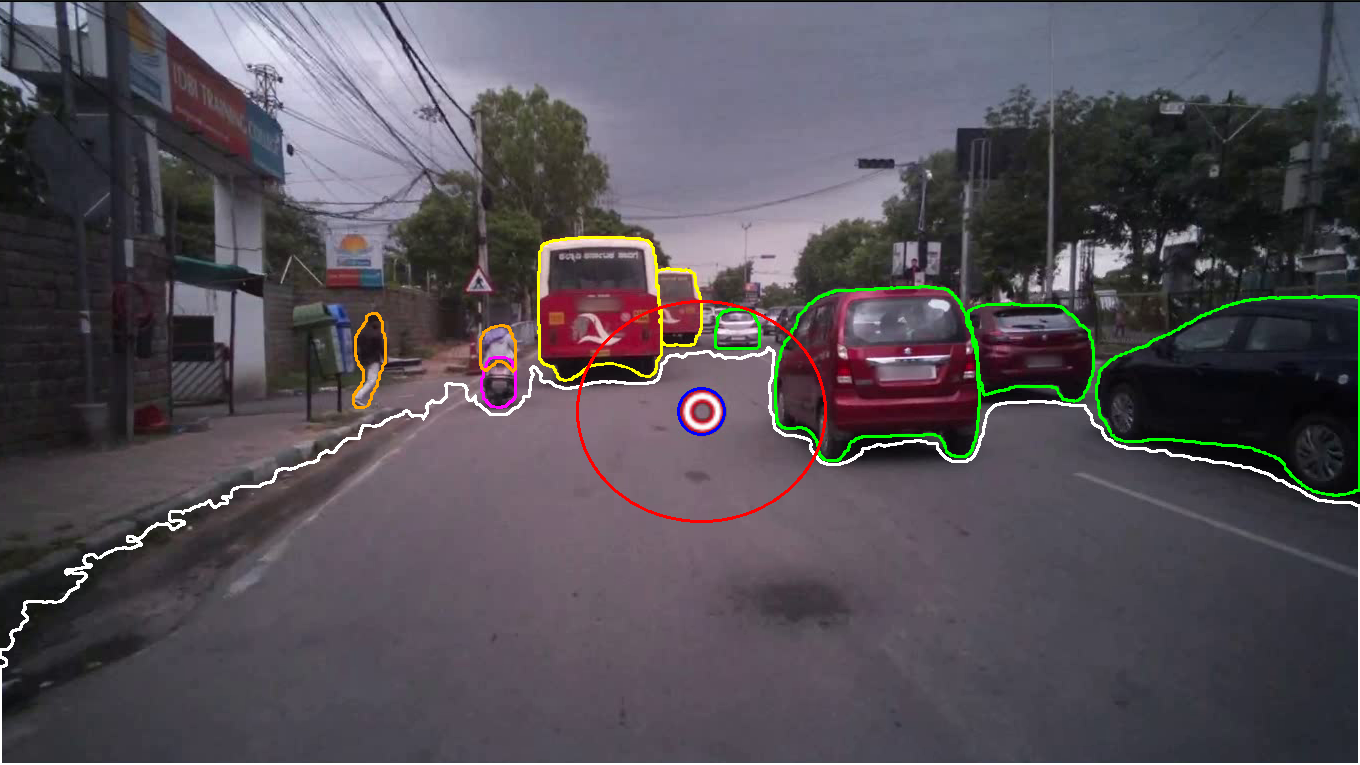}
}
\caption{Example input frame with gaze location (\ref{fig:pipeline_demo}) and corresponding semantic segmentation output with detected objects, road mask, and overlaid central vision region (\ref{fig:pipeline_viz}).}
\label{fig:processing_flow}
\end{figure}

Object detection and tracking were done using YOLOv11x \cite{yolo11_ultralytics} with ByteTrack \cite{zhang2022bytetrack}, chosen for accuracy, speed, and cross-object generalization \cite{yolobenchmark} in handling occlusions and scale variations. Videos were downsampled to 70\% resolution to optimize inference time. Instead of fine-tuning on Indian datasets \cite{paranjape2022dats,varma2019idd}, we focused on fine-tuning the COCO-pretrained YOLOv11x \cite{lin2014microsoft} on underrepresented but frequent Indian traffic objects, specifically autorickshaws \cite{cvit_autorickshaw_2017}, and achieved 92.7\% precision, 75.9\% recall, and 85.9\% mAP@0.5.

Bounding boxes from object detection were passed to SAM2 \cite{ravi2024sam2} for instance segmentation, which supports bounding box prompts allowing seamless integration with YOLOv11x. To address track-level misclassifications, a post-hoc correction was applied by assigning every tracked object its mode class over its track lifespan. Ambiguous cases (without a dominant class) were manually corrected.

For road segmentation, point prompts were generated by selecting pixels for every segmented vehicle located just below the vehicle mask, avoiding any overlap with other objects, and were passed as a point prompt to SAM2 to generate the road mask. Since vehicles typically occupy the road surface, these pixels are likely to belong to the road, allowing the method to produce accurate road masks even under occlusions.

\subsection{Scene-Gaze Mapping and Attention Metrics}

\subsubsection{Attention Attribution: Rider-to-Motorcycle Mapping}

Drivers perceive pedestrians as slow-moving, vulnerable, and unpredictable road users, due to no walkways or zebra crossing. In contrast, motorcyclists are cognitively processed as a unified object with their vehicle, moving cohesively with traffic, a perceptual grouping driven by the principle of common fate \cite{Wagemans2012}.
However, YOLOv11x doesn't explicitly distinguish between pedestrians and motorcyclists, classifying both under a general "person" label, which risks misinterpreting attention patterns. To address this, a post hoc geometric matching spatially maps detected persons to their associated motorcycles, and reclassifies them as motorcyclists. All gaze fixations on a rider are then attributed to the combined rider-vehicle unit, labeled as a new object class: Motorcycle (WA), where WA denotes “with attribution”. 

\subsubsection{Semantic Gaze Mapping and Object-Based Attention Metrics}

Human vision is organized into zones of decreasing acuity and attentional precision, commonly divided into \textit{foveal vision} (0--2\textdegree), \textit{central vision} (0--13\textdegree), \textit{near peripheral} (13--30\textdegree), and \textit{far peripheral} (30--60\textdegree) regions. This study focuses its analysis on two attention categories: \textit{direct gaze} (DG), defined as gaze point overlap with an object, and \textit{central vision} (CV), defined as object presence within a 13\textdegree{} elliptical zone centered at the gaze point, as shown in Figure~\ref{fig:pipeline_viz}.

For characterizing object-wise gaze behavior, in addition to calculating the percentage of frames an object receives direct or central attention (\textit{\%Gaze}) and the probability of receiving attention given its presence in the frame (\textit{P(Gaze)}), as summarized in Table~\ref{tab:gaze_allocation_summary}, the transition probability of gaze between object classes was calculated as:
\begin{equation}
P_{\text{o}}(C_i \rightarrow C_j) = \frac{N_{C_i \rightarrow C_j}}{\sum\limits_{k} N_{C_i \rightarrow C_k}} 
\label{eq:object_transition}
\end{equation}
where $N_{C_i \rightarrow C_j}$ denotes the number of gaze transitions from object class $C_i$ to $C_j$. 

\begin{table}[!htbp]
\centering
\small
\setlength{\tabcolsep}{5pt}
\renewcommand\arraystretch{0.95}
\caption{GAZE ALLOCATION SUMMARY (After Filtering Outliers)}
\begin{tabular}{l c c c}
\toprule
\textbf{Object} & \textbf{\#Obj} & \textbf{\%Gaze} & \textbf{P(Gaze)} \\
\midrule
\multicolumn{4}{l}{\textit{Direct Gaze}}\\
Car            & 1125 / 562  & 8.5 / 7.6   & 20.4 / 17.6 \\
Motorcycle     & 528 / 369   & 3.1 / 2.5   & 16.1 / 7.8  \\
Motorcyclist   & 615 / 283   & 4.8 / 2.4   & 21.5 / 13.6 \\
M/C (WA)       & 153 / 87    & 7.6 / 4.6   & 20.4 / 16.4 \\
Person         & 293 / 161   & 1.2 / 0.9   & 7.0 / 4.4   \\
Autorickshaw   & 385 / 148   & 5.0 / 3.9   & 9.0 / 7.5   \\
Bus            & 70 / 30     & 0.8 / 0.5   & 11.3 / 5.7  \\
Truck          & 156 / 48    & 1.3 / 0.5   & 12.4 / 6.4  \\
Road           & 15 / 10     & 24.2 / 32.4 & 49.4 / 63.6 \\
\midrule
\multicolumn{4}{l}{\textit{Central Vision}}\\
Car            & 2754 / 2166 & 22.9 / 23.8 & 54.9 / 55.2 \\
Motorcycle     & 1937 / 1790 & 15.3 / 15.4 & 51.9 / 45.7 \\
Motorcyclist   & 1580 / 1211 & 15.5 / 13.8 & 54.0 / 52.2 \\
M/C (WA)       & 1137 / 924  & 19.3 / 18.1 & 54.0 / 56.7 \\
Person         & 874 / 927   & 4.5 / 5.5   & 26.7 / 27.5 \\
Autorickshaw   & 653 / 396   & 13.1 / 13.6 & 44.5 / 46.6 \\
Bus            & 128 / 110   & 2.3 / 2.2   & 33.6 / 24.3 \\
Truck          & 312 / 166   & 4.1 / 3.2   & 40.9 / 37.1 \\
Road           & 15 / 10     & 41.3 / 47.2 & 84.3 / 92.5 \\
\bottomrule
\end{tabular}
\label{tab:gaze_allocation_summary}
\vspace{2pt}\\
\textit{Note.} Each cell presents \textbf{experienced}\,/\,\textbf{novice} values.  
\#Obj is the number of uniquely tracked object instances per group.
\end{table}

\subsubsection{Tracked Object Attention Metrics}

When attending to objects in the scene, drivers appear to apply attention to objects in the scene in fragmented bursts, where they quickly glance at the object periodically. To capture this episodic nature of attention, this study defines four temporal metrics for each tracked object, computed independently for direct gaze (DG) and central vision (CV). Let \( o_i \) denote the \( i^\text{th} \) tracked object and \( g \in \{\text{DG}, \text{CV}\} \) represent the attention type. The following terms are defined for each object:

\begin{enumerate}
     \item \( F_i \): The set of frames in which \( o_i \) is continuously tracked. The total number of such frames is \( |F_i| \), and the corresponding duration is given by \( T_i = \frac{|F_i|}{\mathrm{FR}} \), where \( \mathrm{FR} \) is the frame rate.
    
    \item \( A_i^g \subseteq F_i \): The attention set--the subset of frames during which \( o_i \) receives attention of type \( g \).
    
    \item \( S_i^g = \{s_1^g, \dots, s_m^g\} \): The set of $m$ temporally contiguous attention episodes of type \( g \), where each \( s_k^g \subseteq A_i^g \) represents a continuous sequence of attended frames.
\end{enumerate}

The metrics are defined as follows:
\begin{enumerate}
    \item \textbf{Attention Coverage}: Proportion of the object's tracked lifespan during which it receives attention of type \( g \).
    \[
    \mathcal{C}_i^g = \frac{|A_i^g|}{|F_i|}
    \]
    \item \textbf{Episode Rate}: Number of discrete attention episodes per second.
    \[
    \mathcal{R}_i^g = \frac{|S_i^g|}{T_i}
    \]
    \item \textbf{Longest Episode Ratio}: Fraction of the object’s total lifespan occupied by the longest continuous episode.
    \[
    \mathcal{L}_i^g = \frac{\max_{s \in S_i^g} |s|}{|F_i|}
    \]
    \item \textbf{Attention Fragmentation Index}: Degree of fragmentation in attention, defined as the number of episodes relative to total attended frames.
    \[
    \mathcal{F}_i^g = \frac{|S_i^g|}{|A_i^g|}
    \]
\end{enumerate}

Attention metric distributions were assessed using Shapiro-Wilk test, revealing non-normality (p $<$ 0.05 for all metrics). Consequently, non-parametric tests were used along with multiple comparison corrections using Holm-Bonferroni and Benjamini-Hochberg procedures. Primary analyses included:
\begin{enumerate}
    \item Mann-Whitney U tests for between-group comparisons (experienced vs. novice)
    \item Wilcoxon signed-rank tests for within-group comparisons (DG vs. CV)
    \item Kruskal-Wallis H-tests for object class effects
\end{enumerate}

Additionally, an ordinary least squares (OLS) fixed-effects model was used to analyze metric variability, with attention type and driver experience modeled as fixed effects. The use of a fixed-effects approach was justified by near-zero driver-level random intercept variance, indicating that attention patterns did not show significant clustering effects at the individual driver level.

\vspace{-5pt}
\section{Results}

\subsection{Direct Gaze vs Central Vision Usage}

As shown in Table~\ref{tab:attention_metrics_summary}, Central Vision (CV) displayed higher attention coverage as compared to Direct Gaze (DG) across both driver groups ($W = 7.93\times 10^6, p < 0.01, r=0.512$, 95\% CI $\left[ 0.482, 0.543 \right]$). This effect is stronger in novice drivers ($r = 0.556, p < 0.001$) and was confirmed through OLS modeling ($\beta = -0.83, p < 0.001, f^2 = 0.188$).

The temporal structure of attention also differed significantly between the two mechanisms. CV exhibited higher episode rates ($W = 1.23\times10^7, p < 0.001, r = 0.241$) and maintained longer continuous episodes ($W = 8.92\times10^6, p < 0.001, r = 0.451$, 95\% CI $\left[0.418, 0.484\right]$). Conversely, DG showed consistently higher fragmentation across driver groups ($W = 2.17\times10^7, p < 0.001, r = -0.333$ 95\% CI $\left[ -0.364, -0.302\right]$). Bootstrap validation confirmed the stability of attention-type differences for coverage ($\beta = -0.83$, $p < 0.001$, 95\% CI [$-0.88$, $-0.78$]), longest episode ratio ($\beta = -0.63$, $p < 0.001$, 95\% CI [$-0.67$, $-0.59$]), and fragmentation ($\beta = 0.55$, $p < 0.001$, 95\% CI [$0.52$, $0.58$]). In contrast, the episode-rate difference ($\beta = -0.36$, $p < 0.001$, 95\% CI [$-0.39$, $-0.33$]) did not remain robust under bootstrap resampling, indicating sensitivity to sampling variability. All DG–CV comparisons survived Holm-Bonferroni correction for multiple testing.

\subsection{Experience-Related Effects}

Driver experience primarily affected the temporal structure of attention rather than the overall amount of attention allocated. In Central Vision, experienced drivers demonstrated higher episode rates ($U = 4.25 \times 10^7$, $p < 0.001$, $\beta = -0.21$, 95\% model CI [$-0.24$, $-0.18$]) and greater fragmentation ($U = 4.73 \times 10^7$, $p < 0.001$, $\beta = -0.28$, 95\% model CI [$-0.31$, $-0.25$]). Bootstrap validation confirmed that these differences were stable. Direct Gaze patterns showed more modest experience effects. Fragmentation displayed a bootstrap-stable difference, with experienced drivers exhibiting lower fragmentation ($U = 3.51 \times 10^6$, $p < 0.001$, $\beta = -0.28$, 95\% model CI [$-0.30$, $-0.25$]). However, differences in coverage and longest episode ratio were not robust under bootstrap resampling.

Novice drivers exhibited distinct temporal patterns in Central Vision, maintaining higher coverage ($U = 3.20 \times 10^7$, $p < 0.001$, $\beta = 0.25$, 95\% model CI [$0.22$, $0.28$]) and longer continuous episodes ($U = 3.14 \times 10^7$, $p < 0.001$, $\beta = 0.29$, 95\% model CI [$0.26$, $0.31$]) with both effects proved as bootstrap-stable. OLS modeling revealed significant interaction effects between driver experience and attention type for coverage ($\beta = -0.16$, $p < 0.001$) and longest episode ratio ($\beta = -0.21$, $p < 0.001$), although these interactions remained small in magnitude (Cohen's $f^2 < 0.15$).

\begin{figure}[!t]
\centering
\subfloat[]{\includegraphics[width=0.95\linewidth]{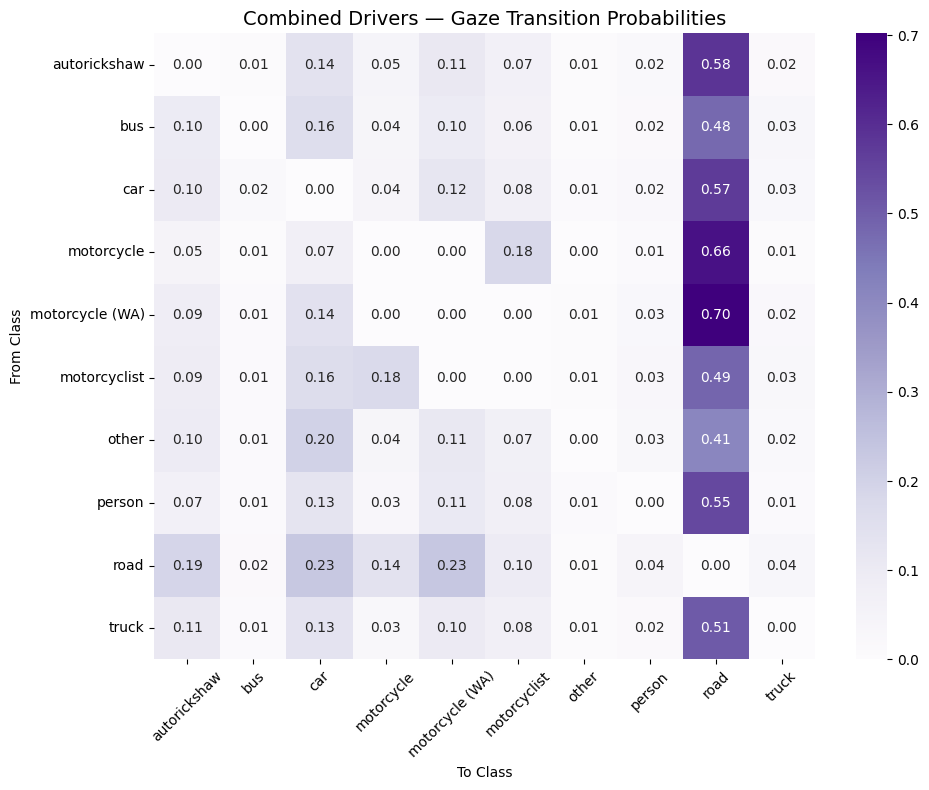}\label{fig:combined_matrix}}\\
\subfloat[]{\includegraphics[width=0.95\linewidth]{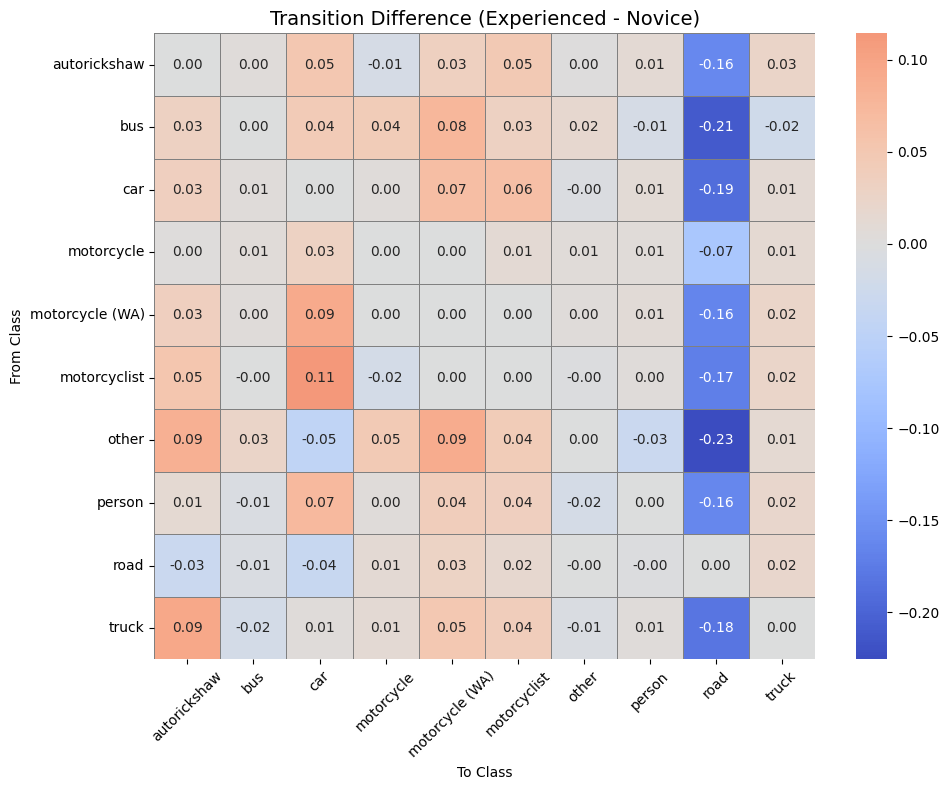}\label{fig:diff_matrix}}
\caption{Gaze transition matrices showing (\ref{fig:combined_matrix}) overall probabilities across all drivers (darker cells indicate higher transition likelihood) and (\ref{fig:diff_matrix}) experience-based differences (warm colors: higher in experienced; cool colors: higher in novice riders).}
\label{fig:transition_matrices}
\end{figure}

\subsection{Object Class Effects}

Object class influence varied significantly by driver experience and attention type. As shown in Table~\ref{tab:gaze_allocation_summary} cars were the most frequent class, followed by motorcycles and motorcyclists. In Direct Gaze, novice drivers exhibited the strongest class effects, with longest-episode ratios showing medium-sized class dependence ($H = 151.48$, $p < 0.001$, $\epsilon^2 = 0.082$) and episode rates also varying substantially across object classes ($H = 122.13$, $p < 0.001$, $\epsilon^2 = 0.066$). Both effects survived Holm-Bonferroni correction, and bootstrap resampling confirmed stability for the longest-episode ratio (bootstrap-confirmed medium effect).

Experienced drivers' Direct Gaze showed smaller class effects, with longest-episode ratio variation ($H = 196.83$, $p < 0.001$, $\epsilon^2 = 0.056$) reaching only the lower medium range, while coverage and fragmentation effects remained small ($\epsilon^2 = 0.027$ and $\epsilon^2 = 0.021$, respectively). Central Vision patterns showed markedly reduced class dependence: experienced drivers’ CV coverage showed negligible class effects ($\epsilon^2 = 0.007$), and even novices’ CV coverage effects remained small ($\epsilon^2 = 0.014$). 

OLS modeling confirmed an interaction between driver experience and attention type for object class modulation, with a significant but small interaction term ($\beta = -0.24$, $p < 0.001$) for longest-episode ratios. Bootstrap validation indicated that only Direct Gaze metrics in novice drivers exhibited medium, reliable object-class effects. All other effects were small in magnitude and, even when statistically significant, did not remain robust under bootstrap validation. 

\section{Discussion}





\subsection{Central Vision and Direct Gaze: Distinct Attention Characteristics}

\begin{table*}[!htbp]
\centering
\small
\setlength{\tabcolsep}{3pt}
\renewcommand\arraystretch{1.05}
\caption{Tracked object attention metrics summary (after outlier filtering)}
\begin{tabular}{l c c c c}
\toprule
\textbf{Object} &
\textbf{Coverage (\%)} &
\textbf{Episode Rate (s$^{-1}$)} &
\textbf{Longest Episode (\%)} &
\textbf{Fragmentation} \\
\midrule
\multicolumn{5}{l}{\textit{Direct Gaze}}\\
Car            & 31.0(13.7) / 34.0(14.6) & 1.6(1.4) / 1.4(1.3) & 26.5(27.3) / 28.5(29.2) & 0.28(0.24) / 0.22(0.21) \\
Motorcycle     & 26.4(12.1) / 34.2(14.5) & 1.7(1.5) / 1.7(1.5) & 21.5(22.4) / 28.9(27.9) & 0.33(0.26) / 0.27(0.24) \\
Motorcyclist   & 26.6(11.6) / 24.9(11.5) & 1.4(1.2) / 1.1(1.0) & 21.3(22.5) / 20.4(22.6) & 0.29(0.25) / 0.24(0.19) \\
M/C (WA)       & 26.9(9.3)  / 28.9(10.9) & 1.3(1.1) / 1.0(0.7) & 16.8(16.9) / 20.3(20.7) & 0.24(0.19) / 0.18(0.12) \\
Person         & 40.8(16.2) / 41.8(16.7) & 2.4(1.7) / 2.0(1.6) & 37.3(32.0) / 38.2(33.4) & 0.34(0.28) / 0.30(0.28) \\
Autorickshaw   & 25.4(10.2) / 21.8(9.5)  & 1.4(1.5) / 1.0(1.1) & 16.6(18.7) / 14.1(17.4) & 0.28(0.24) / 0.22(0.18) \\
Bus            & 32.7(14.5) / 35.6(15.4) & 1.5(1.4) / 1.5(1.2) & 25.7(28.2) / 28.1(32.0) & 0.24(0.19) / 0.23(0.23) \\
Truck          & 35.1(28.7) / 31.6(31.0) & 2.0(1.6) / 1.4(1.5) & 26.2(27.6) / 25.9(31.5) & 0.30(0.23) / 0.26(0.25) \\
Road           & 52.5(13.4) / 63.9(14.5) & 1.2(0.5) / 0.9(0.2) & 3.3(2.6)  / 3.1(1.8)  & 0.10(0.04) / 0.06(0.02) \\
\midrule
\multicolumn{5}{l}{\textit{Central Vision}}\\
Car            & 58.7(16.2) / 66.9(16.5) & 2.2(1.7) / 1.8(1.6) & 47.5(32.9) / 56.9(34.9) & 0.19(0.17) / 0.13(0.13) \\
Motorcycle     & 57.7(16.4) / 70.3(15.8) & 2.2(1.6) / 2.0(1.5) & 47.0(33.0) / 61.1(33.6) & 0.20(0.19) / 0.14(0.14) \\
Motorcyclist   & 55.8(15.6) / 62.6(16.1) & 2.1(1.5) / 1.7(1.3) & 42.8(30.7) / 50.5(32.6) & 0.19(0.18) / 0.14(0.13) \\
M/C (WA)       & 60.6(14.3) / 67.7(14.1) & 2.0(1.5) / 1.6(1.2) & 42.1(29.2) / 50.1(30.7) & 0.16(0.13) / 0.11(0.09) \\
Person         & 62.3(16.5) / 67.8(16.5) & 2.6(1.7) / 2.3(1.5) & 55.3(33.3) / 61.4(34.1) & 0.22(0.19) / 0.17(0.16) \\
Autorickshaw   & 51.7(14.4) / 54.8(14.9) & 2.4(5.8) / 2.4(8.8)& 33.9(27.1) / 38.0(30.5) & 0.20(0.35) / 0.18(0.44) \\
Bus            & 54.6(15.0) / 58.4(17.3) & 1.9(1.5) / 1.7(1.3) & 41.4(30.2) / 47.6(36.3) & 0.16(0.12) / 0.14(0.14) \\
Truck          & 61.6(31.6) / 65.5(33.1) & 2.4(1.8) / 1.7(1.3) & 46.6(32.5) / 54.4(34.9) & 0.20(0.17) / 0.13(0.13) \\
Road           & 84.0(10.5) / 92.3(3.9)  & 1.1(0.6) / 0.7(0.2) & 6.0(4.6)  / 6.9(6.2)  & 0.05(0.03) / 0.03(0.01) \\
\bottomrule
\end{tabular}
\label{tab:attention_metrics_summary}
\vspace{2pt}\\
\textit{Note:} Each cell gives \textbf{experienced}\,/\,\textbf{novice} values. Metrics are medians with half–inter-quartile-range in parentheses.  
M/C (WA) refers to motorcycle with rider attribution. “Road” rows reflect fixations on the roadway itself.
\vspace{-2em}
\end{table*}

Central Vision (CV) operates as the primary monitoring mechanism during driving as it consistently exhibits higher object coverage and longer episode durations as compared to Direct Gaze (DG), as shown in Table~\ref{tab:attention_metrics_summary}. Direct Gaze episodes across both driver groups were shorter, more fragmented and less frequent with small to medium negative effect sizes, suggesting its use for more selective and time-limited sampling of visual information. The stability of these differences across bootstrap analyses and the larger variance explained by attention type compared to experience effects validates the interpretation that these allocation patterns are stable features of driving behavior. Gaze transition patterns also revealed that gaze transitions from viewing an object to the road in over half of all transitions across all driver groups, indicating that the road functions as a default anchor in the drivers' visual scan cycle, as seen in Figure~\ref{fig:combined_matrix}. This could imply that road conditions, like pot-holes is also an attentional metric.

\subsection{Experience-Related Refinements}

Driver experience primarily affects the temporal structure of attention episodes rather than the overall allocation strategy. Experienced drivers exhibited higher episode rates and greater fragmentation in Central Vision indicating faster and more frequent shifts between targets. In Direct Gaze, differences were weaker, with only fragmentation showing a consistent bootstrap-confirmed reduction in experienced drivers, indicating more stable and deliberate foveal probes. Transition patterns reinforced these findings: novices frequently reset gaze to the road after each Direct Gaze episode, while experienced drivers formed more distributed object-to-object transitions across vehicles and vulnerable users, as seen in Figure~\ref{fig:diff_matrix}. The findings suggest that experience affects not just how long drivers attend to objects, but also how they sequence these observations or 'connect the dots', with experienced drivers showing more sophisticated chains of attention between traffic-relevant objects. Although interaction terms between attention type and experience reached statistical significance, their small effect sizes confirmed that experience refines sampling rhythm without altering the core allocation structure.

\subsection{Object-Class Influences}

Object class influenced gaze behavior primarily in novice drivers during Direct Gaze episodes. Variation across classes was most pronounced for the longest-episode ratio, with a medium effect size while other metrics showed smaller effects. Novices tended to dwell longer on cars and motorcycles, reflecting salience-driven attention patterns.  Experienced drivers exhibited some class-dependent variation in Direct Gaze, particularly for longest-episode ratio, but the magnitude of effects was consistently smaller than in novices.  This gradual flattening of class dependence with experience suggests that novice drivers modulate their Direct Gaze behavior moderately based on object identity, while with experience, the class-dependent modulation diminishes, particularly in peripheral monitoring. Central Vision patterns for both groups showed negligible class-dependent variation.

\subsection{Limitations and Future Works}

All recordings were captured in fully natural conditions which improved ecological validity but removed experimental control. 
While the sample size of 25 participants limits broader generalizability, bootstrap validation confirms the robustness of the observed attention pattern. Frequent occlusions and unconstrained participant head movement also produce noise which led to frequent tracking inconsistencies. Future work can integrate more robust object tracking systems that account for factors such as head movement to obtain more robust findings. Finally, the study also assumed Central Vision and Direct Gaze attentions were paired but many foveal fixations arise as promotions from peripheral monitoring, driven by motion of objects or saliency cues, and future work can model this transition to better understand experience-driven attention patterns.

\section{Conclusion}

This study successfully provides the first large-scale analysis of naturalistic gaze behavior on Indian two-wheeler drivers. It presents a semantic segmentation and gaze analysis framework for extracting detailed attention patterns from complex naturalistic data collected in urban, heterogeneous traffic environments.  The findings demonstrate a functional division in gaze behavior: Central Vision operates as a continuous monitoring system, while Direct Gaze supports targeted, time-limited sampling. Novice riders exhibit road-centric gaze behavior, evaluating objects in isolation by returning their gaze to the road after each observation. With experience, this pattern evolves into broader, more distributed chains of attention across road objects. Driver experience also primarily affects the temporal structure of attention episodes rather than the overall allocation strategy. This refinement flattens the influence of object class on gaze allocation, suggesting a shift from salience-driven attention. Collectively, these insights advance the understanding of gaze behavior  optimized through driving experience in complex, real-world settings.


\begin{thebibliography}{00}
 
\bibitem{DataForIndia2024} A. Waghmare, ``Vehicle ownership in India,'' Data For India, Sep. 2024. [Online]. Available: \url{https://www.dataforindia.com/vehicle-ownership} (accessed Oct. 2024).
 
\bibitem{kumar2025myeye2wheeler} B. V. Kumar, D. Rawat, T. Kandalla, A. Nagariya, and K. Vemuri, ``myEye2Wheeler: A Two-Wheeler Indian Driver Real-World Eye-Tracking Dataset,'' in \textit{Proc. IEEE 27th Int. Conf. Intell. Transp. Syst. (ITSC)}, Edmonton, AB, Canada, 2024, pp. 3773--3778, doi: 10.1109/ITSC58415.2024.10920038.
 
\bibitem{wigum2023analysis} J. P. Wigum, P. H. Bogfjellmo, I. Roche-Cerasi, and D. Moe, ``Analysis of Risk Factors in Motorcycle Riding and Distribution of Attention Using Eye Tracking, Interview, and Video---Preliminary Study,'' in \textit{Proc. 33rd Eur. Safety Rel. Conf. (ESREL)}, Southampton, UK, 2023, pp. 675--682, doi: 10.3850/978-981-18-8071-1\_P085-cd.
 
\bibitem{DiStasi2011} L. L. Di Stasi, D. Contreras, A. Candido, J. J. Ca\~{n}as, and A. Catena, ``Behavioral and eye-movement measures to track improvements in driving skills of vulnerable road users: First-time motorcycle riders,'' \textit{Transp. Res. Part F: Traffic Psychol. Behav.}, vol. 14, no. 1, pp. 26--35, 2011.
 
\bibitem{Papakostopoulos2020} V. Papakostopoulos, D. Nathanael, and L. Psarakis, ``Semantic content of motorcycle riders' eye fixations during lane-splitting,'' \textit{Cogn. Technol. Work}, vol. 22, no. 2, pp. 343--355, 2020.
 
\bibitem{aupetit2012naturalistic} S. Aupetit, J. Riff, O. Buttelli, and S. Espi\'{e}, ``Naturalistic study of rider's behaviour in initial training in France: Evidence of limitations in the educational content,'' \textit{Accid. Anal. Prev.}, vol. 58, pp. 206--217, 2013.
 
\bibitem{deniaud2015presence} C. Deniaud, V. Honnet, B. Jeanne, and D. Mestre, ``The concept of `presence' as a measure of ecological validity in driving simulators,'' \textit{J. Interact. Sci.}, vol. 3, Art. no. 1, 2015, doi: 10.1186/s40166-015-0005-z.
 
\bibitem{Kotseruba2021} I. Kotseruba and J. K. Tsotsos, ``Behavioral research and practical models of drivers' attention,'' arXiv:2104.05677, 2021. [Online]. Available: \url{https://arxiv.org/abs/2104.05677}
 
\bibitem{Portela2024} C. Y. Arias-Portela, J. Mora-Vargas, and M. Caro, ``Situational Awareness Assessment of Drivers Boosted by Eye-Tracking Metrics: A Literature Review,'' \textit{Appl. Sci.}, vol. 14, no. 4, Art. no. 1611, 2024, doi: 10.3390/app14041611.
 
\bibitem{Ahlstrom2021} C. Ahlstr\"{o}m, K. Kircher, M. Nystr\"{o}m, and B. Wolfe, ``Eye tracking in driver attention research---How gaze data interpretations influence what we learn,'' \textit{Front. Neuroergonomics}, vol. 2, Art. no. 778043, 2021, doi: 10.3389/fnrgo.2021.778043.
 
\bibitem{holmqvist2011} K. Holmqvist, M. Nystr\"{o}m, R. Andersson, R. Dewhurst, H. Jarodzka, and J. van de Weijer, \textit{Eye Tracking: A Comprehensive Guide to Methods and Measures}. Oxford, U.K.: Oxford Univ. Press, 2011.
 
\bibitem{Palazzi2019} A. Palazzi, D. Abati, S. Calderara, F. Solera, and R. Cucchiara, ``Predicting the driver's focus of attention: The DR(eye)VE project,'' \textit{IEEE Trans. Pattern Anal. Mach. Intell.}, vol. 41, no. 7, pp. 1720--1733, 2019, doi: 10.1109/TPAMI.2018.2845370.
 
\bibitem{Xia2018} Y. Xia, D. Zhang, J. Kim, K. Nakayama, K. Zipser, and D. Whitney, ``Predicting driver attention in critical situations,'' in \textit{Computer Vision -- ACCV 2018} (Lecture Notes in Computer Science, vol. 11365). Cham, Switzerland: Springer, 2019, pp. 658--674.
 
\bibitem{varma2019idd} G. Varma, A. Subramanian, A. Namboodiri, M. Chandraker, and C. V. Jawahar, ``IDD: A dataset for exploring problems of autonomous navigation in unconstrained environments,'' in \textit{Proc. IEEE Winter Conf. Appl. Comput. Vis. (WACV)}, 2019, pp. 1743--1751.
 
\bibitem{paranjape2022dats} B. A. Paranjape and A. A. Naik, ``DATS\_2022: A versatile Indian dataset for object detection in unstructured traffic conditions,'' \textit{Data Brief}, vol. 43, Art. no. 108470, 2022, doi: 10.1016/j.dib.2022.108470.
 
\bibitem{Jackson2009} L. Jackson, P. Chapman, and D. Crundall, ``What happens next? Predicting other road users' behaviour as a function of driving experience and processing time,'' \textit{Ergonomics}, vol. 52, no. 2, pp. 154--164, 2009, doi: 10.1080/00140130802030714.
 
\bibitem{Underwood2011} G. Underwood, D. Crundall, and P. Chapman, ``Driving simulator validation with hazard perception,'' \textit{Transp. Res. Part F: Traffic Psychol. Behav.}, vol. 14, no. 6, pp. 435--446, 2011, doi: 10.1016/j.trf.2011.04.008.
 
\bibitem{Underwood2003} G. Underwood, P. Chapman, N. Brocklehurst, J. Underwood, and D. Crundall, ``Visual attention while driving: Sequences of eye fixations made by experienced and novice drivers,'' \textit{Ergonomics}, vol. 46, no. 6, pp. 629--646, 2003, doi: 10.1080/0014013031000090116.
 
\bibitem{Konstantopoulos2010} P. Konstantopoulos, P. Chapman, and D. Crundall, ``Driver's visual attention as a function of driving experience and visibility. Using a driving simulator to explore drivers' eye movements in day, night and rain driving,'' \textit{Accid. Anal. Prev.}, vol. 42, no. 3, pp. 827--834, 2010, doi: 10.1016/j.aap.2009.09.022.
 
\bibitem{WOLFE2017} B. Wolfe, J. Dobres, R. Rosenholtz, and B. Reimer, ``More than the useful field: Considering peripheral vision in driving,'' \textit{Appl. Ergonom.}, vol. 65, pp. 316--325, 2017, doi: 10.1016/j.apergo.2017.07.009.
 
\bibitem{Papakostopoulos2010} V. Papakostopoulos, D. Nathanael, and N. Marmaras, ``An explorative study of visual scanning strategies of motorcyclists in urban environment,'' in \textit{Proc. 28th Annu. Eur. Conf. Cogn. Ergonom. (ECCE)}, 2010, pp. 157--160, doi: 10.1145/1962300.1962332.
 
\bibitem{yolo11_ultralytics} G. Jocher and J. Qiu, ``Ultralytics YOLO11,'' version 11.0.0, 2024. [Online]. Available: \url{https://github.com/ultralytics/ultralytics}
 
\bibitem{zhang2022bytetrack} Y. Zhang, P. Sun, Y. Jiang, D. Yu, F. Weng, Z. Yuan, P. Luo, W. Liu, and X. Wang, ``ByteTrack: Multi-object tracking by associating every detection box,'' in \textit{Proc. Eur. Conf. Comput. Vis. (ECCV)}, 2022, pp. 1--21.
 
\bibitem{yolobenchmark} N. Jegham, C. Y. Koh, M. Abdelatti, and A. Hendawi, ``Evaluating the evolution of YOLO (You Only Look Once) models: A comprehensive benchmark study of YOLO11 and its predecessors,'' arXiv:2411.00201v1, 2024.
 
\bibitem{lin2014microsoft} T.-Y. Lin, M. Maire, S. Belongie, J. Hays, P. Perona, D. Ramanan, P. Doll\'{a}r, and C. L. Zitnick, ``Microsoft COCO: Common objects in context,'' in \textit{Computer Vision -- ECCV 2014} (Lecture Notes in Computer Science, vol. 8693). Cham, Switzerland: Springer, 2014, pp. 740--755, doi: 10.1007/978-3-319-10602-1\_48.
 
\bibitem{cvit_autorickshaw_2017} CVIT, IIIT Hyderabad, ``Auto-rickshaw detection challenge dataset,'' NCVPRIPG, 2017. [Online]. Available: \url{https://cvit.iiit.ac.in/autorickshaw_detection/}
 
\bibitem{ravi2024sam2} N. Ravi, V. Gabeur, Y.-T. Hu, R. Hu, C. Ryali, T. Ma, H. Khedr, R. R\"{a}dle, C. Rolland, L. Gustafson, E. Mintun, J. Pan, K. V. Alwala, N. Carion, C.-Y. Wu, R. Girshick, P. Doll\'{a}r, and C. Feichtenhofer, ``SAM 2: Segment anything in images and videos,'' in \textit{Proc. Int. Conf. Learn. Represent. (ICLR)}, 2025.
 
\bibitem{Wagemans2012} J. Wagemans, J. H. Elder, M. Kubovy, S. E. Palmer, M. A. Peterson, M. Singh, and R. von der Heydt, ``A century of Gestalt psychology in visual perception: I. Perceptual grouping and figure--ground organization,'' \textit{Psychol. Bull.}, vol. 138, no. 6, pp. 1172--1217, 2012, doi: 10.1037/a0029333.
 
\end{thebibliography}
\end{document}